\documentclass[11pt]{article}

\usepackage[preprint]{acl}

\usepackage{times}
\usepackage{latexsym}
\usepackage{amsmath}
\usepackage{amssymb}

\usepackage[T1]{fontenc}
\usepackage[utf8]{inputenc}

\usepackage{microtype}
\usepackage{inconsolata}
\usepackage{graphicx}

\usepackage{booktabs}
\usepackage{multirow}
\usepackage{siunitx}
\usepackage{array}
\usepackage{tabularx}
\usepackage{adjustbox}
\usepackage{float}

\title{A Mechanistic Investigation of Supervised Fine Tuning}

\author{Ruhaan Chopra \\
  Independent Researcher\\
  \texttt{ruhaanchopra2005@gmail.com}}

\begin{document}
\maketitle
\begin{abstract}
The cosine similarity between a large language model's hidden activations before and after Supervised Fine-Tuning (SFT) remains very high. This, at first glance, suggests that SFT leaves the model’s activation geometry largely undisturbed. However, projecting both sets of activations through a Sparse Autoencoder (SAE) pretrained on the base model reveals that the underlying sparse latents diverge significantly. We introduce a novel investigative pipeline which utilizes these pretrained SAEs as a high-resolution diagnostic tool to mechanistically investigate the drivers of this representational divergence. Through our analytical pipeline, we discover task-specific and layer-specific distributions of the precise semantic features that are systematically altered during supervised fine-tuning. We additionally identify a layer-wise update profile specific to safety alignment. All code, experimental scripts, and analysis files associated with this work are publicly available at: \url{https://github.com/ruhzi/sae-investigation}.

\end{abstract}

\section{Introduction}

Supervised Fine-Tuning (SFT) is the most prevalent immediate post-training intervention used to adapt pretrained large language models for specialized tasks, dialogue formatting, and safety alignment \citep{ouyang2022instructgpt}.

A naive analysis of the activation space suggests that SFT barely perturbs the model's internal geometry. Empirically, we compare hidden states before and after fine-tuning across four independently fine-tuned variants of Gemma 3 1B, where each variant is initialized from the same base model and trained on a single isolated task. Across all evaluated tasks and layers, the raw activation cosine similarity between the base and fine-tuned models remains high ($> 0.96$). This raises the mechanistic question: if the raw activation vectors remain nearly identical in direction, what underlying structures are actually being rewritten?

To examine this, we introduce a novel investigative pipeline utilizing Sparse Autoencoders (SAEs) \citep{sae} as a high-resolution diagnostic tool. Pretrained SAEs reconstruct dense activation vectors as sparse linear combinations of interpretable feature directions, effectively acting as a magnifying glass for the localized semantic subspaces of the model. By projecting both the pre-SFT and post-SFT activations into a static SAE dictionary trained on the base model, we use the resulting sparse latents as a proxy to quantify representational drift. 

Our investigation reveals a steep, continuous decline in the cosine similarity of SAE latents as fine-tuning progresses, a degradation that diverges sharply from the stable raw activations. This aligns with previous observations regarding the fragility of a pretrained SAE's ability to reconstruct activations following fine-tuning \citep{kissane2024saetransfer}. This representational drift is consistent across varying SAE dictionary widths (16k to 262k) and gradient checkpoints, but it exhibits striking layer-wise heterogeneity. For example, during semantic reasoning fine-tuning (\textsc{MultiNLI}), the raw activation cosine similarity remains rigidly high at both early and late layers ($0.996$ at Layer 7; $0.960$ at Layer 22). However, the SAE latent cosine similarity exposes a severe degradation gradient: dropping to $0.900$ at Layer 7, decaying to $0.847$ at Layer 13, and plummeting to $0.509$ by Layer 22. Similar layer-wise trajectories are observed across safety (\textsc{WildJailbreak}) and mathematical reasoning (\textsc{GSM8K}) tasks, where the latent similarity steadily decays at each optimization step. This localized divergence provides empirical evidence that SFT attempts to rewrite narrow semantic subspaces surgically. Our next objective is to find out if it succeeds.

We isolate the representational shift by computing the activation difference matrix,
\[
\Delta H = H_{\text{SFT}} - H_{\text{base}},
\]
across our evaluation sets. We then apply Singular Value Decomposition (SVD) to $\Delta H$ to extract the principal components (PCs) that capture the maximum variance of the fine-tuning update.

We subsequently compute the cosine similarity between the dominant PCs of $\Delta H$ and the individual feature vectors within the static SAE decoder dictionary. Because the dictionary serves as a topological map of the base model's internal geometry, PCs that exhibit extreme alignment (either highly positive or strongly negative cosine similarity) reveal the precise semantic directions being amplified or suppressed by the SFT gradients.

We additionally inject the top principal components into the base hidden states and pass these modified activations through the pretrained SAE. As discussed previously, the base SAE provides a frozen semantic basis and therefore acts as a strict structural baseline for isolating the impact of the update. This forward pass allows us to identify exactly which sparse features are forced to flip between active and inactive states, from which we compute the average flip rate to quantify the disruption across different semantic clusters.

We programmatically retrieve precomputed, human-interpretable feature descriptions associated with the most disrupted features (i.e., features exhibiting extreme PC--dictionary alignment and high interventional flip rates) via the Neuronpedia API \citep{api}. Through qualitative review of these explanations, we derive a discrete semantic taxonomy consisting of seven functional clusters: \emph{Persona}, \emph{Structure}, \emph{Code}, \emph{Reasoning}, \emph{Safety}, \emph{Multilingual}, and \emph{Collateral} (unrelated domain-specific trivia). We then implement an automated LLM-as-a-judge annotation pipeline utilizing GPT-5.4-mini to classify all features \citep{judge}.

This pipeline helps uncover a task and layer-specific profile of subspace rewrites during SFT. We also discover that safety alignment disproportionately affects earlier transformer layers, in stark contrast to every other task. This points to safety alignment being unlike every other behavioral injection, and one which gets shaped very early on. Finally, we also reveal that, contrary to the notion inferred from observing a seemingly undisturbed activation space post-SFT, fine-tuning may not be fully surgical: we observe unrelated features consistently disrupted across diverse objectives.
\section{Preliminaries}

\subsection{Supervised Fine-Tuning}

Supervised fine-tuning (SFT) is a post-pretraining adaptation technique whereby a pretrained language model is further optimised on a curated dataset of input--output pairs using standard supervised learning objectives. Given a base model pretrained on large-scale corpora via next-token prediction, SFT adapts the model's parameters toward a target distribution of desired behaviors---such as instruction-following, question answering, or domain-specific generation---by minimising the cross-entropy loss over labeled demonstration data \citep{llama}.

Formally, let
\[
\mathcal{D} = \{(x_i, y_i)\}_{i=1}^{N}
\]
denote a dataset of prompt--response pairs. SFT optimises the conditional language modelling objective
\[
\mathcal{L}_{\text{SFT}}
=
-\sum_{i=1}^{N}
\log p_{\theta}(y_i \mid x_i),
\]
where $\theta$ denotes the model parameters.

SFT is widely employed as the first stage of alignment pipelines, often preceding reinforcement learning from human feedback (RLHF), and has been shown to be a powerful mechanism for instilling task-specific and behaviorally-specific competencies into general-purpose language models.

\subsection{Sparse Autoencoders and GemmaScope 2}

A central challenge in mechanistic interpretability is the phenomenon of \emph{polysemanticity}, wherein individual neurons activate across semantically unrelated contexts, resisting human-interpretable explanation. A hypothesised cause of polysemanticity is \emph{superposition}, whereby models encode more features than the dimensionality of the layer permits by representing them as an overcomplete set of non-orthogonal directions in activation space.

Sparse autoencoders (SAEs) have been proposed as a scalable, unsupervised approach to resolving superposition by learning a dictionary of sparsely-activating directions that decompose internal activations into more monosemantic components \citep{sae2}.

Formally, an SAE maps an activation vector
\[
\mathbf{x} \in \mathbb{R}^{d}
\]
to a high-dimensional sparse representation via an encoder
\[
\mathbf{c}
=
\mathrm{ReLU}(\mathbf{M}\mathbf{x} + \mathbf{b}),
\]
and reconstructs it via a decoder
\[
\hat{\mathbf{x}}
=
\mathbf{M}^{\top}\mathbf{c},
\]
where
\[
\mathbf{M} \in \mathbb{R}^{d_{\text{hid}} \times d}
\]
constitutes the learned feature dictionary with
\[
d_{\text{hid}} \gg d.
\]

Training minimises the composite objective
\[
\mathcal{L}
=
\|\mathbf{x} - \hat{\mathbf{x}}\|_2^2
+
\alpha \|\mathbf{c}\|_1,
\]
balancing reconstruction fidelity against sparsity of feature activations. The resulting dictionary features have been shown to be substantially more interpretable and monosemantic than neuron-basis directions or alternative decompositions such as PCA and ICA.

Building upon this foundation, \emph{GemmaScope 2} \citep{scope} was released by Google as a comprehensive open-source suite of interpretability tools covering the full Gemma 3 model family \citep{gemma}, ranging from 270M to 27B parameters. GemmaScope 2 combines sparse autoencoders and transcoders to allow researchers to examine models' internal representations and trace how these representations connect to observed model behavior.

\section{Methodology}

\subsection{Overview}

We investigate the effect of supervised fine-tuning (SFT) on the internal representations of a pretrained language model, using sparse autoencoders (SAEs) as a diagnostic lens. Our methodology proceeds in four stages:
\begin{enumerate}
    \item we fine-tune a base model independently on each of four structurally diverse tasks;
    \item we measure representational drift at both the raw activation level and the SAE latent level throughout training;
    \item we perform a principal component analysis of the activation drift to identify the dominant directions of representational change; and
    \item we probe each drift direction through the pretrained SAE to identify the specific latent features disrupted by fine-tuning, characterizing them via automated interpretability.
\end{enumerate}

\subsection{Model and Sparse Autoencoders}

\paragraph{Base model.}
We use Gemma 3 1B IT (\texttt{google/gemma-3-1b-it}), an instruction-tuned transformer with model dimension
\[
d_{\text{model}} = 1152
\]
and 26 layers. All experiments use \texttt{bfloat16} precision.

\paragraph{Sparse autoencoders.}
We use the publicly available Gemma Scope 2 SAEs (\texttt{google/gemma-scope-2-1b-it}), trained on residual stream activations of the base model.

For each of three probe layers---layer 7 (early), layer 13 (middle), and layer 22 (late)---we load SAEs at three dictionary widths:
\[
16\text{k} \quad (d_{\text{SAE}} = 16{,}384),
\]
\[
65\text{k} \quad (d_{\text{SAE}} = 65{,}536),
\]
\[
262\text{k} \quad (d_{\text{SAE}} = 262{,}144),
\]
all using the medium $L_0$ sparsity setting. This yields nine SAE configurations in total.

The SAEs employ a JumpReLU activation function with learned threshold $\tau$:
\[
\mathbf{z}
=
\mathrm{JumpReLU}_{\tau}
\big(
(\mathbf{h} - \mathbf{b}_{\mathrm{dec}})
\mathbf{W}_{\mathrm{enc}}
+
\mathbf{b}_{\mathrm{enc}}
\big),
\]
where:
\begin{itemize}
    \item $\mathbf{h} \in \mathbb{R}^{d_{\text{model}}}$ is the residual stream activation,
    \item $\mathbf{W}_{\mathrm{enc}} \in \mathbb{R}^{d_{\text{model}} \times d_{\text{SAE}}}$,
    \item $z_i = 0$ whenever the pre-activation falls below the learned threshold $\tau_i$.
\end{itemize}

The decoder matrix
\[
\mathbf{W}_{\mathrm{dec}}
\in
\mathbb{R}^{d_{\text{SAE}} \times d_{\text{model}}}
\]
reconstructs the activation as
\[
\hat{\mathbf{h}}
=
\mathbf{z}\mathbf{W}_{\mathrm{dec}}
+
\mathbf{b}_{\mathrm{dec}}.
\]

Critically, SAE weights are frozen throughout all experiments. The SAEs serve purely as a diagnostic instrument: since they were trained to faithfully decompose the original model's activations into interpretable features, any change in latent outputs after fine-tuning reflects a genuine representational shift rather than adaptation of the SAE itself.

\subsection{Tasks and Datasets}

To characterize how fine-tuning objectives influence representational drift, we select four structurally distinct tasks.

\begin{table*}[t]
\centering
\small
\begin{tabular}{l l l c c}
\hline
\textbf{Task} & \textbf{Dataset} & \textbf{Domain} & \textbf{$N_{\text{train}}$} & \textbf{$N_{\text{test}}$} \\
\hline
Natural Language Inference & \textsc{MultiNLI \citep{multinli}} & Logical reasoning & 2000 & 100 \\
Mathematical Reasoning & \textsc{GSM8K \citep{gsm8k}} & Chain-of-thought arithmetic & 2000 & 100 \\
Safety Alignment & \textsc{WildJailbreak} & Refusal of harmful requests & 2000 & 100 \\
Structured Tool Calling & OpenAI Tool Calling \citep{tool} & JSON function invocation & 50 & 13 \\
\hline
\end{tabular}
\caption{Fine-tuning tasks used in the study.}
\label{tab:tasks}
\end{table*}

For \textsc{WildJailbreak} \citep{jail}, we filter to the \texttt{vanilla\_harmful} subset. For tool calling, we flatten the structured message format into a single training string. All datasets are shuffled using random seed
\[
s = 42.
\]

\subsection{Fine-Tuning Protocol}

For each task, we perform full-parameter supervised fine-tuning starting from a fresh copy of the base model.

\paragraph{Training details.}
\begin{itemize}
    \item Optimizer: Adafactor
    \item Learning rate:
    \[
    \eta = 2 \times 10^{-5}
    \]
    \item No learning-rate warmup
    \item No relative step scaling
    \item Gradient accumulation: 4
    \item Gradient clipping: 1.0
\end{itemize}

\subsection{Activation and Latent Cosine Similarity Tracking}

We define a held-out evaluation set with
\[
N_{\text{test}} = 100
\]
examples per task.

Before training begins, we extract baseline activations:
\[
H_l^{(0)}
=
\{
\mathbf{h}^{(0)}_{(l,j)}
\}_{j=1}^{N_{\text{test}}}
\]
from residual stream layer
\[
l \in \{7, 13, 22\}.
\]

We also compute baseline SAE latents:
\[
Z^{(0)}_{(l,w)}
\]
for each SAE configuration.

Every 400 training samples, we pause training and recompute activations and SAE latents on the same evaluation set.

We define activation cosine similarity as
\[
\mathrm{CosSim}_{\mathrm{act}}(l,t)
=
\frac{1}{N_{\text{test}}}
\sum_{j}
\frac{1}{T_j}
\sum_{p}
\frac{
\mathbf{h}^{(0)}_{(l,j,p)}
\cdot
\mathbf{h}^{(t)}_{(l,j,p)}
}{
\|
\mathbf{h}^{(0)}_{(l,j,p)}
\|
\,
\|
\mathbf{h}^{(t)}_{(l,j,p)}
\|
},
\]
where:
\begin{itemize}
    \item $T_j$ is sequence length,
    \item $p$ indexes token positions.
\end{itemize}

We analogously compute latent cosine similarity:
\[
\mathrm{CosSim}_{\mathrm{SAE}}(l,w,t).
\]

\subsection{Subspace Drift Analysis via SVD}

After training, we construct the drift matrix:
\[
\Delta_l
=
F_l - V_l
\in
\mathbb{R}^{N \times d_{\text{model}}},
\]
where:
\begin{itemize}
    \item $V_l$ denotes baseline activations,
    \item $F_l$ denotes fine-tuned activations,
    \item $N$ is the total number of token positions.
\end{itemize}

We center the drift matrix:
\[
\widetilde{\Delta}_l
=
\Delta_l - \mathrm{mean}(\Delta_l).
\]

When
\[
N > 2000,
\]
we uniformly subsample to 2000 tokens.

We then compute truncated singular value decomposition:
\[
\widetilde{\Delta}_l
=
U \Sigma V^{\top},
\]
where:
\begin{itemize}
    \item $\Sigma = \mathrm{diag}(\sigma_1, \sigma_2, \dots)$,
    \item rows of $V^{\top}$ are principal drift directions $\mathbf{v}_i$.
\end{itemize}

Variance explained by component $i$ is:
\[
\frac{\sigma_i^2}{\sum_k \sigma_k^2}.
\]

We retain the smallest $k$ such that cumulative variance explained reaches $90\%$, capped at
\[
k_{\max} = 50.
\]

Formally:
\[
k
=
\min
\left(
k_{\max},
\min
\left\{
k'
:
\frac{
\sum_{i=1}^{k'}
\sigma_i^2
}{
\sum_j
\sigma_j^2
}
\geq 0.90
\right\}
\right).
\]

\subsection{Perturbation Probing and Feature Flip Analysis}

For each drift direction $\mathbf{v}_i$, we perturb baseline activations:
\[
\widetilde{\mathbf{h}}_j
=
\mathbf{h}_j
+
\epsilon_i \mathbf{v}_i,
\]
where perturbation magnitude is:
\[
\epsilon_i
=
\frac{\sigma_i}{\sqrt{N}}.
\]

This corresponds to a one-standard-deviation perturbation along the $i$-th principal drift direction.

We apply perturbations to
\[
M = 500
\]
baseline token activations.

A feature flip occurs when feature $f$ changes activation state:
\[
\mathrm{flip}_{(i,f,j)}
=
\mathbf{1}
\left[
\mathbf{1}[z_f(\mathbf{h}_j) > 0]
\neq
\mathbf{1}[z_f(\widetilde{\mathbf{h}}_j) > 0]
\right].
\]

We aggregate flips in two ways.

\paragraph{Flip rate per direction}
\[
F_i
=
\frac{1}{M}
\sum_j
\sum_f
\mathrm{flip}_{(i,f,j)}.
\]

\paragraph{Per-feature flip frequency}
\[
\phi_{(i,f)}
=
\frac{1}{M}
\sum_j
\mathrm{flip}_{(i,f,j)}.
\]

\subsection{Identification of Outlier Directions}

We identify ``outlier directions'' using a z-score criterion:
\[
z_i
=
\frac{
F_i - \mathrm{mean}(F)
}{
\mathrm{std}(F) + 10^{-8}
}.
\]

Directions satisfying
\[
z_i > 1.5
\]
are flagged as outliers.

If no directions exceed threshold, we instead select the top five directions by raw flip count.

For each outlier direction, we compute:
\begin{itemize}

    \item cosine similarity:
    \[
    \cos(\mathbf{w}^{\mathrm{dec}}_f, \mathbf{v}_i),
    \]
    \item exact flip frequency:
    \[
    \phi_{(i,f)}.
    \]
\end{itemize}

\subsection{Semantic Characterization via Automated Interpretability}

To interpret feature indices semantically, we query the neuronpedia API, which provides natural-language explanations for Gemma Scope SAE features.

For each outlier direction, we retrieve explanations for:
\begin{itemize}

    \item top-10 features by cosine similarity,
    \item top-10 features by flip frequency,
\end{itemize}
yielding up to 20 characterized features per direction.

This provides a semantic characterization of which linguistic capabilities or knowledge domains are most disrupted by fine-tuning along each principal drift axis.

\subsection{Functional Taxonomy via LLM-as-a-Judge}

We classify each disrupted feature into a seven-category taxonomy:
\begin{enumerate}
    \item \emph{Persona} (conversational formatting, greetings, hedging),
    \item \emph{Structure} (markdown, bullet points, JSON scaffolding),
    \item \emph{Code} (programming constructs, API artifacts, documentation),
    \item \emph{Reasoning} (logical connectives, mathematical steps, causal markers),
    \item \emph{Safety} (refusal phrasing, legal disclaimers, harm-related content),
    \item \emph{Multilingual} (foreign-language tokens, script suppression),
    \item \emph{Collateral} (hyper-specific or functionally unrelated features).
\end{enumerate}

This taxonomy was developed inductively from manual inspection of feature descriptions.

We assign each feature to a category using an LLM-as-a-judge approach. Each feature description is presented to GPT-5.4-mini with a fixed system prompt (Appendix~\ref{appendix:system prompt}) specifying the taxonomy definitions and requiring structured JSON output containing:
\begin{itemize}
    \item the predicted cluster,
    \item a confidence score in $[0,1]$,
    \item a one-sentence mechanistic justification.
\end{itemize}

We process all features from both the aligned and flipped feature reports. The resulting annotations enable us to compute per-task, per-layer distributions over functional categories, revealing which behavioral dimensions of SFT are most responsible for the observed representational drift.

\section{Findings and Discussion}

\setlength{\tabcolsep}{2pt}
\renewcommand{\arraystretch}{0.68}

%%%%%%%%%%%%%%%%%%%%%%%%%%%%%%%%%%%%%%%%%%%%%%%%%%%%%%%%%%%%
% TABLE 1
%%%%%%%%%%%%%%%%%%%%%%%%%%%%%%%%%%%%%%%%%%%%%%%%%%%%%%%%%%%%

\begin{table}[!t]
\centering
\tiny
\caption{
Cosine similarity between hidden activations before and after SFT across training steps.
Even after substantial fine-tuning, raw activation vectors remain highly aligned.
}
\label{tab:activation-cossim}

\vspace{-2mm}

\begin{tabular}{lcccc}
\toprule
\textbf{Task} & \textbf{Step} & \textbf{Layer 7} & \textbf{Layer 13} & \textbf{Layer 22} \\
\midrule
\multirow{5}{*}{MultiNLI}
& 400  & 0.997 & 0.993 & 0.971 \\
& 800  & 0.997 & 0.993 & 0.966 \\
& 1200 & 0.996 & 0.992 & 0.963 \\
& 1600 & 0.996 & 0.991 & 0.962 \\
& 2000 & 0.996 & 0.992 & 0.960 \\
\midrule
\multirow{5}{*}{GSM8K}
& 400  & 0.998 & 0.995 & 0.975 \\
& 800  & 0.997 & 0.994 & 0.973 \\
& 1200 & 0.996 & 0.994 & 0.970 \\
& 1600 & 0.996 & 0.993 & 0.969 \\
& 2000 & 0.996 & 0.993 & 0.968 \\
\midrule
\multirow{5}{*}{WildJailbreak}
& 400  & 0.999 & 0.997 & 0.984 \\
& 800  & 0.999 & 0.997 & 0.983 \\
& 1200 & 0.998 & 0.996 & 0.982 \\
& 1600 & 0.998 & 0.996 & 0.981 \\
& 2000 & 0.998 & 0.996 & 0.980 \\
\midrule
ToolCalling
& 50 & 0.999 & 0.995 & 0.976 \\
\bottomrule
\end{tabular}

\vspace{-4mm}
\end{table}

%%%%%%%%%%%%%%%%%%%%%%%%%%%%%%%%%%%%%%%%%%%%%%%%%%%%%%%%%%%%
% TABLE 2
%%%%%%%%%%%%%%%%%%%%%%%%%%%%%%%%%%%%%%%%%%%%%%%%%%%%%%%%%%%%

\begin{table}[!t]
\centering
\tiny
\caption{
Cosine similarity between SAE(262k) latent representations before and after SFT.
In contrast to raw activations, sparse latent representations diverge substantially during fine-tuning,
particularly in deeper layers.
}
\label{tab:sae-cossim}

\vspace{-2mm}

\begin{tabular}{lcccc}
\toprule
\textbf{Task} & \textbf{Step} & \textbf{Layer 7} & \textbf{Layer 13} & \textbf{Layer 22} \\
\midrule
\multirow{5}{*}{MultiNLI}
& 400  & 0.909 & 0.924 & 0.740 \\
& 800  & 0.900 & 0.896 & 0.678 \\
& 1200 & 0.893 & 0.872 & 0.637 \\
& 1600 & 0.883 & 0.837 & 0.602 \\
& 2000 & 0.874 & 0.830 & 0.557 \\
\midrule
\multirow{5}{*}{GSM8K}
& 400  & 0.927 & 0.934 & 0.793 \\
& 800  & 0.914 & 0.926 & 0.781 \\
& 1200 & 0.907 & 0.916 & 0.753 \\
& 1600 & 0.902 & 0.920 & 0.718 \\
& 2000 & 0.894 & 0.909 & 0.708 \\
\midrule
\multirow{5}{*}{WildJailbreak}
& 400  & 0.932 & 0.914 & 0.762 \\
& 800  & 0.919 & 0.903 & 0.746 \\
& 1200 & 0.907 & 0.890 & 0.725 \\
& 1600 & 0.901 & 0.888 & 0.716 \\
& 2000 & 0.898 & 0.883 & 0.707 \\
\midrule
ToolCalling
& 50 & 0.949 & 0.957 & 0.873 \\
\bottomrule
\end{tabular}

\vspace{-4mm}
\end{table}

%%%%%%%%%%%%%%%%%%%%%%%%%%%%%%%%%%%%%%%%%%%%%%%%%%%%%%%%%%%%
% TABLE 3
%%%%%%%%%%%%%%%%%%%%%%%%%%%%%%%%%%%%%%%%%%%%%%%%%%%%%%%%%%%%

\begin{table}[!t]
\centering
\tiny
\caption{
Number of principal-component directions whose induced SAE feature flips are statistically anomalous relative to other activation-delta directions.
}
\label{tab:cursed-directions}

\vspace{-2mm}

\resizebox{\columnwidth}{!}{
\begin{tabular}{llccc}
\toprule
\textbf{Dataset} & \textbf{Layer} & \textbf{SAE Config} & \textbf{\# Directions} & \textbf{Max Variance (\%)} \\
\midrule

\multirow{6}{*}{SFT MultiNLI}
& Layer 7  & L7\_16k   & 1 & 78.69 \\
& Layer 7  & L7\_65k   & 1 & 78.69 \\
& Layer 7  & L7\_262k  & 1 & 78.69 \\
& Layer 22 & L22\_16k  & 1 & 32.59 \\
& Layer 22 & L22\_65k  & 1 & 32.59 \\
& Layer 22 & L22\_262k & 1 & 32.59 \\
\midrule

\multirow{6}{*}{SFT GSM8K}
& Layer 7  & L7\_16k   & 1 & 67.95 \\
& Layer 7  & L7\_65k   & 1 & 67.95 \\
& Layer 7  & L7\_262k  & 1 & 67.95 \\
& Layer 22 & L22\_16k  & 3 & 17.23 \\
& Layer 22 & L22\_65k  & 3 & 17.23 \\
& Layer 22 & L22\_262k & 2 & 17.23 \\
\midrule

\multirow{6}{*}{SFT WildJailbreak}
& Layer 7  & L7\_16k   & 3 & 89.29 \\
& Layer 7  & L7\_65k   & 3 & 89.29 \\
& Layer 7  & L7\_262k  & 3 & 89.29 \\
& Layer 22 & L22\_16k  & 1 & 20.32 \\
& Layer 22 & L22\_65k  & 1 & 20.32 \\
& Layer 22 & L22\_262k & 1 & 20.32 \\
\midrule

\multirow{6}{*}{SFT ToolCalling}
& Layer 7  & L7\_16k   & 1 & 74.77 \\
& Layer 7  & L7\_65k   & 1 & 74.77 \\
& Layer 7  & L7\_262k  & 1 & 74.77 \\
& Layer 22 & L22\_16k  & 4 & 17.73 \\
& Layer 22 & L22\_65k  & 5 & 17.73 \\
& Layer 22 & L22\_262k & 5 & 17.73 \\

\bottomrule
\end{tabular}
}

\vspace{-4mm}
\end{table}

\subsection{Cosine Similarity} The raw activation vectors remain highly stable during fine-tuning, maintaining a cosine similarity above 0.960 throughout training (Table~\ref{tab:activation-cossim}). For example, WildJailbreak Layer 22 shifts only minimally from 0.984 to 0.980 over the course of training. In stark contrast, projecting those same activations into a pretrained SAE reveals massive hidden divergence, with latent cosine similarity dropping as low as 0.557 (Table~\ref{tab:sae-cossim}). The representational rewriting worsens significantly in deeper layers, as seen in MultiNLI (Step 2000) where Layer 7 degrades to 0.874, Layer 13 to 0.830, and Layer 22 plummets to 0.557 (Table~\ref{tab:sae-cossim}). This degradation happens continuously at every training step, demonstrated by GSM8K Layer 22 decaying steadily from 0.793 at step 400 down to 0.708 at step 2000 (Table~\ref{tab:sae-cossim}). Despite these dramatic latent shifts, the corresponding raw activation vectors for the same layers remain highly aligned across training (Table~\ref{tab:activation-cossim}). This phenomenon is consistently observed across all four fine-tuning tasks (MultiNLI, GSM8K, WildJailbreak, ToolCalling) and persists across all three tested SAE dictionary widths (16k, 65k, and 262k).
\begin{figure*}[t]
    \centering
    \includegraphics[width=\textwidth]{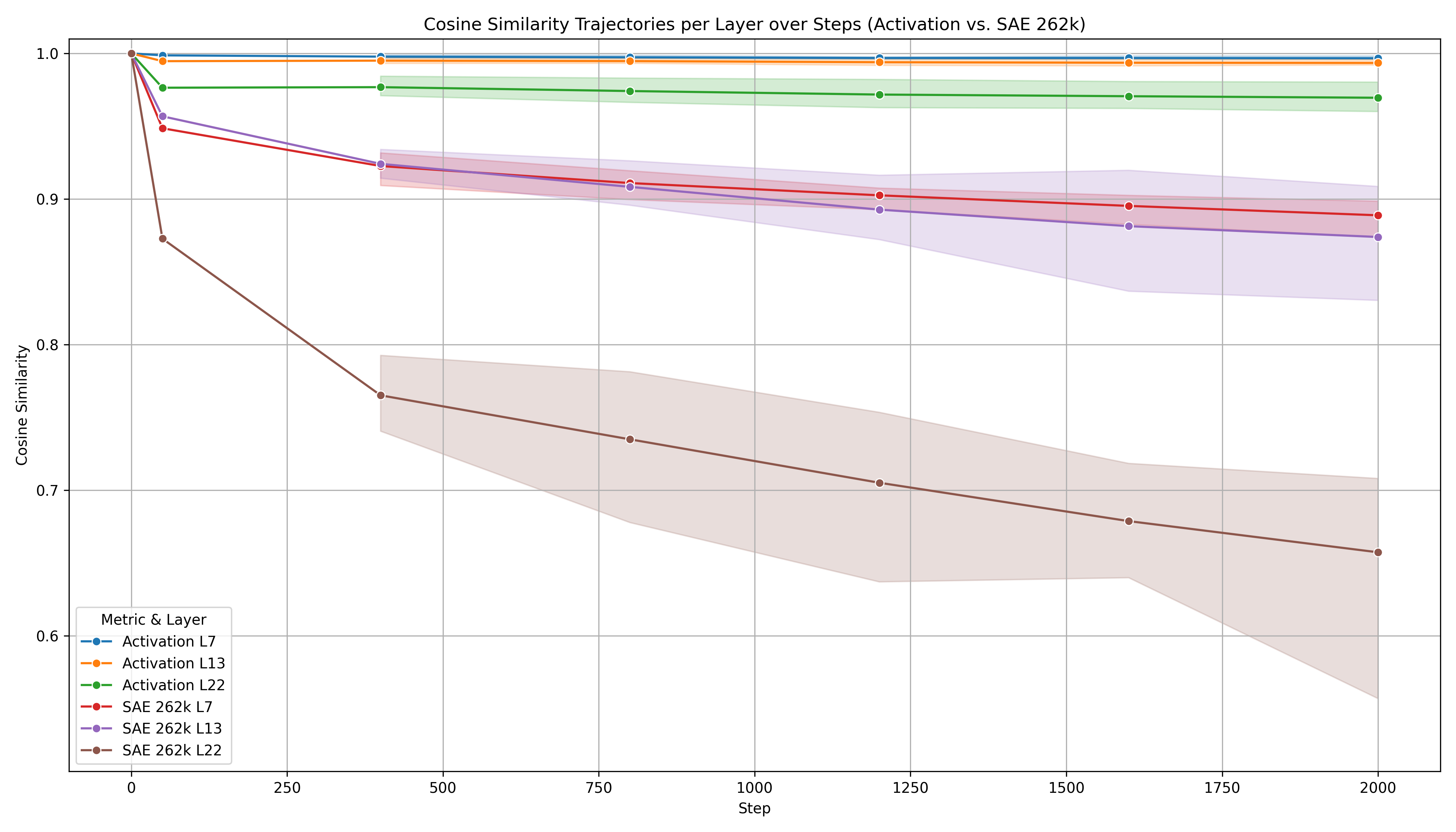}
    \caption{
    Cosine similarity trajectories between pretrained and fine-tuned representations across training steps.
    Raw hidden activations remain highly aligned throughout SFT, while SAE latent representations diverge substantially, particularly in deeper layers.
    }
    \label{fig:cossim-trajectories}
\end{figure*}

% Requires:
% \usepackage{booktabs}
% \usepackage{multirow}
% \usepackage{siunitx}

% Recommended setup in preamble:
%
% \sisetup{
%   round-mode          = places,
%   round-precision     = 3,
%   table-number-alignment = center
% }

%%%%%%%%%%%%%%%%%%%%%%%%%%%%%%%%%%%%%%%%%%%%%%%%%%%%%%%%%%%%
% TABLE 4
%%%%%%%%%%%%%%%%%%%%%%%%%%%%%%%%%%%%%%%%%%%%%%%%%%%%%%%%%%%%

\begin{table}[H]
\centering
\tiny
\caption{
Percentage distribution of flipped SAE features across semantic clusters after SFT.
Percentages are computed independently within each experiment-layer pair.
}
\label{tab:feature-flips}

\vspace{-2mm}

\resizebox{\columnwidth}{!}{
\begin{tabular}{lllc}
\toprule
\textbf{Experiment} & \textbf{Layer} & \textbf{Semantic Cluster} & \textbf{Flipped Features (\%)} \\
\midrule

\multirow{12}{*}{SFT GSM8K}
& 7  & Collateral     & 50.00 \\
& 7  & Structure      & 10.00 \\
& 7  & Code           & 10.00 \\
& 7  & Reasoning      & 10.00 \\
& 7  & Safety         & 10.00 \\
& 7  & Multilingual   & 10.00 \\
& 22 & Code           & 25.00 \\
& 22 & Structure      & 20.00 \\
& 22 & Persona        & 15.00 \\
& 22 & Reasoning      & 15.00 \\
& 22 & Collateral     & 15.00 \\
& 22 & Multilingual   & 10.00 \\
\midrule

\multirow{9}{*}{SFT MultiNLI}
& 7  & Persona        & 30.00 \\
& 7  & Structure      & 30.00 \\
& 7  & Collateral     & 20.00 \\
& 7  & Code           & 10.00 \\
& 7  & Safety         & 10.00 \\
& 22 & Code           & 40.00 \\
& 22 & Collateral     & 30.00 \\
& 22 & Structure      & 20.00 \\
& 22 & Safety         & 10.00 \\
\midrule

\multirow{12}{*}{SFT ToolCalling}
& 7  & Collateral     & 40.00 \\
& 7  & Code           & 30.00 \\
& 7  & Persona        & 10.00 \\
& 7  & Reasoning      & 10.00 \\
& 7  & Multilingual   & 10.00 \\
& 22 & Code           & 26.00 \\
& 22 & Reasoning      & 18.00 \\
& 22 & Structure      & 14.00 \\
& 22 & Persona        & 12.00 \\
& 22 & Collateral     & 12.00 \\
& 22 & Safety         & 10.00 \\
& 22 & Multilingual   & 8.00 \\
\midrule

\multirow{11}{*}{SFT WildJailbreak}
& 7  & Reasoning      & 20.00 \\
& 7  & Collateral     & 20.00 \\
& 7  & Persona        & 16.67 \\
& 7  & Structure      & 16.67 \\
& 7  & Code           & 10.00 \\
& 7  & Multilingual   & 10.00 \\
& 7  & Safety         & 6.67 \\
& 22 & Reasoning      & 40.00 \\
& 22 & Structure      & 30.00 \\
& 22 & Code           & 20.00 \\
& 22 & Collateral     & 10.00 \\

\bottomrule
\end{tabular}
}

\vspace{-4mm}
\end{table}

\subsection{Subspace Analysis}

A consistent layer-wise shift in the rank structure of the SFT update is observed across tasks (Table~\ref{tab:cursed-directions}). In early layers (Layer 7), the representational rewrite is low-rank and highly concentrated: a single principal component accounts for 78.7\%, 74.8\%, and 68.0\% of activation-delta variance for MultiNLI, ToolCalling, and GSM8K respectively. In deeper layers (Layer 22), this structure becomes substantially more distributed---the variance explained by the leading direction falls to 32.6\% (MultiNLI), 17.7\% (ToolCalling), and 17.2\% (GSM8K). This pattern is consistent across all three SAE dictionary widths, suggesting that early-layer SFT operates via broad, low-dimensional contextual shifts, whereas deeper-layer adaptation requires more distributed, multi-dimensional semantic rerouting.

A prominent cross-task finding is that SFT systematically rewrites features with no direct functional relationship to the training objective. The clearest instance is the pervasive disruption of \textit{Code} features in tasks with no programming component: Code is the single largest disrupted cluster at Layer 22 in both GSM8K (25.0\%) and MultiNLI (40.0\%), despite neither task involving code (Table~\ref{tab:feature-flips}). Similarly, \textit{Collateral} features---those encoding domain-specific content entirely unrelated to the task---dominate shallow-layer disruption across tasks, accounting for 50.0\% of all flipped features at Layer 7 in GSM8K and 40.0\% in ToolCalling (Table~\ref{tab:feature-flips}). These findings suggest that early SFT updates operate through broad, imprecise perturbations of the representation space, and that certain feature clusters, particularly Code-related ones, serve as general-purpose structural scaffolds that fine-tuning repurposes across semantically unrelated objectives. 
\textit{Note:} Results are shown for the 262k SAE width for the reasons discussed in Appendix~\ref{appendix:width}.

A structurally distinct pattern emerges for safety alignment. WildJailbreak is the only task where the ratio of Layer 7 to Layer 22 flipped features exceeds 1, at 3.00 (30 vs.\ 10 features), compared to ratios of 0.50, 1.00, and 0.20 for GSM8K, MultiNLI, and ToolCalling respectively (Table~\ref{tab:layer-flip-ratios}). This inversion suggests that safety-oriented SFT concentrates its representational rewrite in shallower layers to a degree not observed in any other task type. 

Across all tasks and both layers, the Collateral cluster consistently accounts for the largest share of SAE dictionary features whose cosine similarity with the principal outlier direction exceeds $|0.5|$ (Table~\ref{tab:outlier-cossim-features}). This finding indicates that the geometrically dominant axis of the SFT update is oriented primarily toward semantically incidental subspaces, rather than toward features encoding the task-relevant competencies the fine-tuning is designed to instill.

%%%%%%%%%%%%%%%%%%%%%%%%%%%%%%%%%%%%%%%%%%%%%%%%%%%%%%%%%%%%
% TABLE 5
%%%%%%%%%%%%%%%%%%%%%%%%%%%%%%%%%%%%%%%%%%%%%%%%%%%%%%%%%%%%

\begin{table}[!t]
\centering
\tiny
\caption{
Distribution of SAE dictionary features whose cosine similarity with the principal outlier directions exceeds $0.5$ or falls below $-0.5$.
Features are grouped by semantic cluster using their natural-language descriptions.
The final column indicates which principal outlier directions the features align with.
}
\label{tab:outlier-cossim-features}

\vspace{-2mm}

\resizebox{\columnwidth}{!}{
\begin{tabular}{llll}
\toprule
\textbf{Experiment} & \textbf{Layer} & \textbf{Semantic Cluster} & \textbf{Outlier Directions} \\
\midrule

\multirow{8}{*}{SFT GSM8K}
& 7  & Collateral (5)    & [0] \\
& 7  & Structure (2)     & [0] \\
& 7  & Multilingual (2)  & [0] \\
& 7  & Code (1)          & [0] \\
& 22 & Collateral (7)    & [0] \\
& 22 & Persona (1)       & [0] \\
& 22 & Reasoning (1)     & [0] \\
& 22 & Safety (1)        & [0] \\
\midrule

\multirow{8}{*}{SFT MultiNLI}
& 7  & Structure (4)     & [0] \\
& 7  & Collateral (4)    & [0] \\
& 7  & Code (1)          & [0] \\
& 7  & Multilingual (1)  & [0] \\
& 22 & Collateral (6)    & [0] \\
& 22 & Safety (2)        & [0] \\
& 22 & Persona (1)       & [0] \\
& 22 & Reasoning (1)     & [0] \\
\midrule

\multirow{8}{*}{SFT ToolCalling}
& 7  & Structure (4)     & [0] \\
& 7  & Collateral (4)    & [0] \\
& 7  & Code (1)          & [0] \\
& 7  & Multilingual (1)  & [0] \\
& 22 & Collateral (6)    & [0, 3] \\
& 22 & Safety (3)        & [0] \\
& 22 & Persona (1)       & [0] \\
& 22 & Structure (1)     & [0] \\
\midrule

\multirow{11}{*}{SFT WildJailbreak}
& 7  & Reasoning (5)     & [1] \\
& 7  & Structure (4)     & [0] \\
& 7  & Collateral (4)    & [0, 1] \\
& 7  & Persona (3)       & [1] \\
& 7  & Multilingual (3)  & [0, 1] \\
& 7  & Code (1)          & [0] \\
& 22 & Collateral (5)    & [0] \\
& 22 & Safety (2)        & [0] \\
& 22 & Persona (1)       & [0] \\
& 22 & Structure (1)     & [0] \\
& 22 & Reasoning (1)     & [0] \\

\bottomrule
\end{tabular}
}

\vspace{-4mm}
\end{table}

%%%%%%%%%%%%%%%%%%%%%%%%%%%%%%%%%%%%%%%%%%%%%%%%%%%%%%%%%%%%
% TABLE 6
%%%%%%%%%%%%%%%%%%%%%%%%%%%%%%%%%%%%%%%%%%%%%%%%%%%%%%%%%%%%

\begin{table}[!t]
\centering
\tiny
\caption{
Total number of flipped SAE features at shallow and deep layers for each task.
The final column reports the ratio of Layer 7 to Layer 22 flipped features.
}
\label{tab:layer-flip-ratios}

\vspace{-2mm}

\begin{tabular}{lccc}
\toprule
\textbf{Task} & \textbf{Layer 7} & \textbf{Layer 22} & \textbf{L7 / L22 Ratio} \\
\midrule
SFT GSM8K         & 10 & 20 & 0.50 \\
SFT MultiNLI      & 10 & 10 & 1.00 \\
SFT ToolCalling   & 10 & 50 & 0.20 \\
SFT WildJailbreak & 30 & 10 & 3.00 \\
\bottomrule
\end{tabular}

\vspace{-4mm}
\end{table}

\section{Conclusion}

We introduce a novel mechanistic analysis pipeline that leverages pretrained SAEs as a frozen semantic basis. Since these SAEs were trained prior to fine-tuning and therefore are strictly invariant to it, they enable post-hoc attribution of representational change to the fine-tuning process itself. We validate our pipeline by investigating supervised fine-tuning across a structurally diverse set of tasks. We find that SFT leaves raw activation geometry largely undisturbed, yet projects into substantially altered sparse latent representations, revealing that the apparent representational stability of the activation space masks significant underlying semantic drift. We additionally observe that, unlike every other task studied, safety alignment causes most feature-level disruptions to occur in earlier layers. These disruptions correspond to binary state changes in interpretable semantic units that indicate which learned capabilities are being actively reconfigured. No other fine-tuning objective exhibits this structural asymmetry. This points to safety-alignment possibly being a completely different behavioral capability which shapes differently and earlier than other behaviors like semantic reasoning or tool-calling.  Finally, despite the apparent geometric preservation of activation space, SFT is not fully surgical: features such as Code are disrupted in a task-invariant manner across objectives with no explicit programming component, indicating that collateral representational interference is an inherent characteristic of the fine-tuning process rather than a task-specific artifact.

\section*{Limitations}
Our analysis is currently limited to the GemmaScope 2 interpretability suite and a single model scale, Gemma 3 1B IT. As a result, it remains unclear whether the representational patterns observed here generalize across other architectures, parameter scales, or fine-tuning regimes. Additionally, our pipeline is fundamentally constrained by the quality, granularity, and layer coverage of the available SAEs.

\bibliography{custom}
\appendix

\section{System Prompt}
\label{appendix:system prompt}

The following system prompt was used for all automated feature annotations:

\begin{quote}
\small
\begin{verbatim}
Role:
You are an expert AI researcher specializing in
Mechanistic Interpretability. Your task is to
analyze human-readable descriptions of Sparse
Autoencoder (SAE) features extracted from a
Large Language Model and classify them into
one of seven distinct functional clusters.

Context:
These features were identified during a study
on Supervised Fine-Tuning (SFT). SFT alters
the base model's latent space to enforce
structural formatting, safety guardrails,
conversational personas, and specific
reasoning pathways. 
The Taxonomy:
You must classify every feature into exactly
one of the following seven categories:

1. Persona
(The Assistant Persona & Conversational
Formatting): Greetings, apologies, offering
help, "my name is", "please".

2. Structure
(Structural Scaffolding & Syntax):
Bullet points, markdown delimiters, JSON
formatting, list separators.

3. Code
(Code, Tooling, & Technical Artifacts):
Code snippets, Python/JavaScript syntax,
API terminology, HTML tags, technical
documentation.

4. Reasoning
(Reasoning Anchors & Logical Connectors):
"because", "therefore", mathematical
deductions, causal phrasing, chain-of-thought
style reasoning.

5. Safety
(Safety, Guardrails, & Sensitive Content):
Violence, refusal phrasing, legal disclaimers,
harm-related content, moderation behavior.

6. Multilingual
(Multilingual Suppression & Alignment):
Foreign-language conjunctions, multilingual
tokens, Cyrillic/Asian scripts, non-English
suffixes.

7. Collateral
(Collateral Damage / Hyper-Specific Oddities):
Random trivia, hyper-specific entities,
unrelated people/places, semantically
incidental concepts.

Instructions:
1. Analyze the feature description deeply.
2. Determine which behavioral goal of SFT
most likely explains the feature.
3. If the feature appears unrelated to any
assistant capability, default to Collateral.
4. Return output in strict JSON format.

Output Schema:
{
  "predicted_cluster": "<cluster>",
  "confidence_score": <float>,
  "reasoning": "<brief mechanistic justification>"
}
\end{verbatim}
\end{quote}

\section{Width-wise Flip Rate}
\label{appendix:width}

To evaluate whether the observed feature-disruption patterns depend on SAE dictionary width, we repeated the feature-flip analysis across all three GemmaScope 2 SAE configurations (16k, 65k, and 262k) at both shallow (Layer 7) and deep (Layer 22) probe layers.

Table~\ref{tab:collateral-width} summarizes the proportion of flipped features classified as Collateral across all evaluated SAE widths. While the qualitative patterns reported in the main paper remain broadly stable across widths, several width-dependent effects emerge.

At Layer 22, larger SAE dictionaries consistently exhibit a lower proportion of flipped features classified as Collateral. This trend is strongest for ToolCalling and WildJailbreak. For ToolCalling Layer 22, the proportion of Collateral features decreases monotonically from 42.5\% for the 16k SAE to 26.0\% for the 65k SAE and further to 12.0\% for the 262k SAE. WildJailbreak Layer 22 shows a similarly monotonic decline from 40.0\% to 20.0\% to 10.0\%. GSM8K Layer 22 also follows this pattern, decreasing from 36.7\% (16k) and 33.3\% (65k) to 15.0\% at 262k.

The trend is less stable at Layer 7. For example, GSM8K Layer 7 increases from 30.0\% Collateral at 16k to 50.0\% at both 65k and 262k, while MultiNLI Layer 7 fluctuates between 20.0\%, 40.0\%, and 20.0\% across widths. These results suggest that shallow-layer feature organization may remain comparatively unstable across SAE resolutions, whereas deeper-layer disentanglement improves more consistently with larger dictionaries.

Importantly, the core qualitative findings of the paper remain preserved at 262k width, despite the reduction in Collateral prevalence. In particular:
\begin{itemize}
    \item safety alignment continues to exhibit an unusually shallow-layer-heavy flip profile,
    \item deeper layers continue to exhibit substantially larger latent divergence,
    \item and non-task-specific feature disruption remains observable even at the highest SAE resolution.
\end{itemize}

The reduction in Collateral prevalence at larger widths may indicate improved feature disentanglement. Smaller SAEs necessarily compress a larger number of semantic directions into coarser superposed features, which can cause unrelated concepts to appear bundled together under broad ``Collateral'' activations. Wider SAEs provide a substantially larger representational dictionary and may therefore separate semantically distinct concepts more cleanly, reducing the frequency with which fine-tuning perturbations manifest through highly mixed or incidental features.

For this reason, all feature-alignment and feature-flip analyses presented in the main paper use the 262k SAE configuration. The 262k SAEs provide the highest available representational granularity and exhibit the lowest overall prevalence of broad Collateral features in most Layer 22 settings, making them the most conservative and semantically disentangled probe among the evaluated SAE widths.

\begin{table*}[t]
\centering
\footnotesize
\setlength{\tabcolsep}{5pt}
\renewcommand{\arraystretch}{0.95}

\caption{
Relationship between SAE dictionary width and the proportion of flipped features classified as Collateral across tasks and layers.
}
\label{tab:collateral-width}

\begin{tabular}{llccc}
\toprule
\textbf{Experiment} & \textbf{SAE} & \textbf{Total Flipped} & \textbf{Collateral Features} & \textbf{Collateral (\%)} \\
\midrule

SFT GSM8K & L22\_16k  & 30 & 11 & 36.67 \\
SFT GSM8K & L22\_65k  & 30 & 10 & 33.33 \\
SFT GSM8K & L22\_262k & 20 & 3  & 15.00 \\
SFT GSM8K & L7\_16k   & 10 & 3  & 30.00 \\
SFT GSM8K & L7\_65k   & 10 & 5  & 50.00 \\
SFT GSM8K & L7\_262k  & 10 & 5  & 50.00 \\

\midrule

SFT MultiNLI & L22\_16k  & 10 & 3 & 30.00 \\
SFT MultiNLI & L22\_65k  & 10 & 2 & 20.00 \\
SFT MultiNLI & L22\_262k & 10 & 3 & 30.00 \\
SFT MultiNLI & L7\_16k   & 10 & 2 & 20.00 \\
SFT MultiNLI & L7\_65k   & 10 & 4 & 40.00 \\
SFT MultiNLI & L7\_262k  & 10 & 2 & 20.00 \\

\midrule

SFT ToolCalling & L22\_16k  & 40 & 17 & 42.50 \\
SFT ToolCalling & L22\_65k  & 50 & 13 & 26.00 \\
SFT ToolCalling & L22\_262k & 50 & 6  & 12.00 \\
SFT ToolCalling & L7\_16k   & 10 & 7  & 70.00 \\
SFT ToolCalling & L7\_65k   & 10 & 3  & 30.00 \\
SFT ToolCalling & L7\_262k  & 10 & 4  & 40.00 \\

\midrule

SFT WildJailbreak & L22\_16k  & 10 & 4 & 40.00 \\
SFT WildJailbreak & L22\_65k  & 10 & 2 & 20.00 \\
SFT WildJailbreak & L22\_262k & 10 & 1 & 10.00 \\
SFT WildJailbreak & L7\_16k   & 30 & 8 & 26.67 \\
SFT WildJailbreak & L7\_65k   & 30 & 7 & 23.33 \\
SFT WildJailbreak & L7\_262k  & 30 & 6 & 20.00 \\

\bottomrule
\end{tabular}
\end{table*}

\end{document}